%% file: ecai-sample-and-instructions.tex
\documentclass[final]{ecai}  
\paperid{1972}        
\usepackage{amsmath,amsfonts}
\usepackage{algorithm, algpseudocode}
\usepackage{array}
\usepackage[caption=true,font=normalsize,labelfont=sf,textfont=sf]{subfig}
\usepackage{textcomp}
\usepackage{stfloats}
\usepackage{url}
\usepackage{booktabs}
\usepackage{verbatim}
\usepackage{graphicx}
\usepackage{command}

\usepackage{bbm}
\usepackage{latexsym}

\begin{document}

\begin{frontmatter}

\title{LFAA: Crafting Transferable Targeted Adversarial Examples with Low-Frequency Perturbations}

\author[A]{\fnms{Kunyu}~\snm{Wang}\thanks{Corresponding Author. Email: kunyuwang@link.cuhk.edu.hk}}
\author[A]{\fnms{Juluan}~\snm{Shi}}
\author[A]{\fnms{Wenxuan}~\snm{Wang}} 

\address[A]{The Chinese University of Hong Kong}

\begin{abstract}
    Deep neural networks are susceptible to adversarial attacks, which pose a significant threat to their security and reliability in real-world applications. The most notable adversarial attacks are transfer-based attacks, where an adversary crafts an adversarial example to fool one model, which can also fool other models. While previous research has made progress in improving the transferability of untargeted adversarial examples, the generation of targeted adversarial examples that can transfer between models remains a challenging task. In this work, we present a novel approach to generate transferable targeted adversarial examples by exploiting the vulnerability of deep neural networks to perturbations on high-frequency components of images. We observe that replacing the high-frequency component of an image with that of another image can mislead deep models, motivating us to craft perturbations containing high-frequency information to achieve targeted attacks. To this end, we propose a method called Low-Frequency Adversarial Attack (\name), which trains a conditional generator to generate targeted adversarial perturbations that are then added to the low-frequency component of the image. Extensive experiments on ImageNet demonstrate that our proposed approach significantly outperforms state-of-the-art methods, improving targeted attack success rates by a margin from 3.2\% to 15.5\%.
\end{abstract}

\end{frontmatter}

\section{Introduction}
Deep neural networks (DNNs) have achieved remarkable progress in numerous domains~\cite{he2016resnet}. However, the security and dependability of DNNs remain difficult to ensure when confronted with adversarial examples~\cite{goodfellow2015FGSM,szegedy2014intriguing}. Adversarial examples are created with malicious intent and include slight alterations that are imperceptible to the human eye but are sufficient to deceive deep models into making incorrect predictions. This weakness poses a threat to security-sensitive applications such as face recognition.

The study of adversarial attacks can be divided into two categories: white-box attacks and black-box attacks. White-box attacks have access to the victim model's architecture and parameters to create adversarial perturbations. In contrast, black-box attacks have limited or no knowledge of the victim model. Existing white-box attacks use gradient information to generate adversarial examples~\cite{goodfellow2015FGSM,madry2018pgd}. Adversarial examples created on white-box models can deceive unknown neural models~\cite{Papernot2017blackbox}, making black-box attacks feasible. Several methods have been proposed to enhance the transferability of adversarial examples~\cite{wang2021admix,dong2018boosting,lin2020nesterov}, and they have achieved remarkable performance in the untargeted black-box scenario, where they only aim to produce different predictions. However, these methods are not satisfactory in the targeted attack scenario, where adversaries intend to trick the models into making specific predictions. Several approaches have been proposed to improve the transferability of adversarial examples in the targeted setting~\cite{zhao2021success,li2020towards,wang2021Feature,long2022frequency}, with the generator-based method being the most effective. This approach trains a generator based on the source model to produce adversarial examples for the target class~\cite{yang2022boosting,poursaeed2018generative}.

\input{figs/frequency_ex}

Prior research on targeted attacks~\cite{goodfellow2015FGSM,zhao2021success,li2020towards} has focused on generating adversarial perturbations, but these perturbations typically do not contain any information from the target class image, leading to overfitting of the models. To overcome this issue, it is important to modify the texture and shape of the input image to match the characteristics of the target label image. By incorporating such information, adversaries can generate adversarial examples with improved transferability compared to directly adding noise.

The high-frequency component of an image, as depicted in Fig.~\ref{fig:freq}, typically captures the detailed textures and noise, while the low-frequency component represents the object's shape. Since the background of an image is unnecessary for classification and can easily cause overfitting, it is essential to directly manipulate the image frequency domain information to embed the target class image information into the input image, which can improve transferability. With this insight in mind, we proposed a novel generation-based attack called Low-Frequency Adversarial Attack (\name). The method trains a conditional generator to produce adversarial frequency domain information corresponding to the target label, which is then added to the low-frequency component of the input image to generate adversarial examples (See Fig. 2). As depicted in Fig.~\ref{fig:freq}, the perturbation modifies the high-frequency gesture and blurs the low-frequency component to emphasize high-frequency component. As a result, the classifier misclassifies the high-frequency of the adversarial examples as the target label image, indicating that we have embedded the target label information in the frequency component. Specifically, we optimize the generator by minimizing the classification loss between the predictions of the adversarial examples and the target label.

In summary, we highlight our contributions as follows:
\begin{itemize}
\item We demonstrate that modifying the frequency domain information can effectively mislead deep models.
\item We propose a novel attack called Low-Frequency Adversarial Attack (\name), which is the first generation-based targeted attack that generates adversarial frequency information to embed target label information into the source image.
\item Empirical evaluations on the ImageNet dataset demonstrate \name achieves much better transferability than the state-of-the-art target attacks.
\end{itemize}

\section{Related Work}
In this section, we review related adversarial attacks in the black-box setting, frequency-based attacks, and defenses related to our proposed attack.

\textbf{Adversarial Attack} Szegedy \etal were the first to demonstrate the vulnerability of deep neural networks to adversarial examples~\cite{szegedy2014intriguing}. Subsequently, several works have investigated the susceptibility of deep models to black-box attacks, where access to the target model is restricted. Such attacks can be categorized into three types: (a) score-based attacks, which can access the predicted probability ~\cite{ilyas2018black}; (b) decision-based attacks, which can only obtain the predicted label~\cite{brendel2018decision}; and (c) transfer-based attacks, which are effective in real-world settings where the attacker cannot query the target model. In this approach, the adversary crafts the adversarial examples on a white-box model and transfers them to fool the target model~\cite{lin2020nesterov,wang2021admix}. Therefore, the transferability of adversarial examples is crucial for deceiving unknown models.

The Fast Gradient Sign Method (FGSM) was the first gradient-based attack proposed, which generates perturbations in the direction of the gradient~\cite{goodfellow2015FGSM}. The iterative FGSM is an extension of FGSM that produces better white-box performance but worse black-box performance. However, to avoid overfitting to the white-box model, several methods have been proposed to escape local maxima. For instance, MI-FGSM~\cite{dong2018boosting} adds momentum to I-FGSM to stabilize the optimization process. Lin \etal further improve the method by incorporating the Nesterov accelerated gradient, which provides an effective lookahead strategy~\cite{lin2020nesterov}.

In addition to gradient-based attacks, data augmentation techniques have also been found to be effective in improving the transferability of adversarial examples. For instance, the Diverse Input Method (DIM) \cite{xie2019improving} rescales the input image to random sizes and adds padding to a fixed size before calculating the gradient. The Transition Invariant Attack \cite{dong2019evading} approximates the gradient calculation for a set of translated images by convolving the gradient with a Gaussian kernel. Admix \cite{wang2021admix} mixes the input image with a small portion of images from other categories. Some methods also focus on modifying surrogate models to enhance the transferability of iterative methods. For example, skip connections \cite{wu2020skip} have been used to improve the transferability of adversarial examples. Finally, other attacks aim to increase the feature difference between the source image and adversarial examples \cite{wang2021Feature}.

Although the methods mentioned above have shown good performance in untargeted attacks, their performance in targeted attacks is usually poor. To address this, several researchers have proposed advanced loss functions for targeted attacks. Li \etal~\cite{li2020towards} adopt Poincare distance and Triplet loss to replace cross-entropy, which has a vanishing gradient in iterative targeted attacks. Zhao \etal~\cite{zhao2021success} use the logits output of the target class as the loss function and perform a large number of iterations to achieve state-of-the-art performance. Gao \etal~\cite{gao2021feature} minimize the distance between the features of the target sample and adversarial examples in the reproducing kernel Hilbert space, which is transition-invariant. Another branch of attack is generative-based attacks, where attackers train a generator to generate perturbations given an input image. Generative-based attacks are more efficient because they can learn the adversarial pattern of a target label using a large dataset. The adversarial pattern depends on the entire data distribution and is not limited to a single image, which may overfit the source model. UAP \cite{Moosavi2017universal} was the first proposed method to fool models by learning universal noise, GAP \cite{poursaeed2018generative} learns a generator that can produce image-agnostic perturbations for targeted attacks, CDA \cite{naseer2019cross} uses relativistic training objectives to boost cross-domain transferability, and C-GSP \cite{yang2022boosting} uses a conditional generator to generate targeted adversarial perturbations. Our method belongs to the generative-based method, but instead of generating perturbations directly, we generate frequency content that can easily fool deep models. We seek to learn the frequency components of a target class image based on a data distribution.


\textbf{Frequency-based Attack} According to recent studies, researchers have explored the generalization and adversarial vulnerability of deep neural networks (DNNs) from a frequency perspective \cite{yin2019fourier,wang2020high,sharma2019on}. These studies indicate that DNNs can capture high-frequency components that are imperceptible to humans. Yin \etal~ \cite{yin2019fourier} demonstrate that naturally trained models are vulnerable to perturbations on high-frequency components, while adversarially trained models are less sensitive to such perturbations. As a result, several approaches have been proposed to craft adversarial examples from the perspective of frequency. For example, Long \etal~\cite{long2022frequency} perturb input images with Gaussian noise in the frequency domain as a data augmentation technique, then convert them back to the spatial domain for gradient calculation to enhance transferability. Guo \etal~\cite{guo2019low} restrict the search space to the low-frequency domain to craft adversarial examples, showing that low-frequency components are important in model decision making for query-based attacks. Sharma \etal~\cite{sharma2019on} craft adversarial examples by randomly masking low-frequency components and demonstrate that even adversarially trained models are still vulnerable to low-frequency perturbations. Finally, Zhang \etal~ \cite{zhang2022practical} propose a method of crafting adversarial examples by replacing the high-frequency portion of an image with a handcrafted adversarial patch. However, this method cannot be used for targeted attacks, and the selection of the adversarial patch remains a concern.

\textbf{Adversarial Defense}
As the threat of adversarial attacks on deep neural networks (DNNs) continues to increase, various defense methods have been proposed to mitigate this problem. Adversarial training, which involves injecting adversarial examples into the training process, is a promising method that has shown success in improving model robustness \cite{goodfellow2015FGSM,madry2018pgd}. Tram{\`e}r~\etal \cite{tramer2018ensemble} introduced ensemble adversarial training, which uses adversarial examples generated on multiple models to improve the robustness of the resulting model. Another approach to adversarial defense involves using denoising filters, which remove strange patterns from adversarial examples before feeding them to the classifier. For example, Liao \etal~\cite{liao2018defense} proposed a High-level representation guided denoiser (HGD) to suppress perturbations, while Naseer \etal~\cite{naseer2020nrp} trained a neural representation purifier (NRP) that learns to purify perturbed input images. Other defense methods utilize input transformations to mitigate the effects of adversarial perturbations, such as random resizing and padding (R\&P) \cite{xie2019improving}, and feature distillation \cite{liu2019FD}.
\section{Method}
\input{figs/method}
In this section, we will present the adversarial setup and provide a comprehensive description of our proposed \name.

\subsection{Preliminaries}
\label{method:pre}
We begin by considering an image $\mathbf{x}\in \mathbf{R}^{H\times W\times C}$ from a dataset $\mathcal{X}$. An adversarial example $\mathbf{x}_{adv}$ is generated such that it is nearly identical to the original image, i.e., $||\mathbf{x} - \mathbf{x}_{adv}||_p \le \epsilon$, where $||||_p$ is the $L_{p}$ norm distance and $\epsilon$ is the perturbation budget. We adopt the $L_{\infty}$ distance metric in this work. We consider a classification neural network $\mathcal{F}$ with parameters $\phi$ and loss function $\mathbf{J}$ that is trained to classify images into a set of classes $\mathcal{C} = {c_1,c_2,\dots,c_d}$. The function $\mathcal{F}_{\theta}: \mathbf{R}^{H\times W\times C} \rightarrow \mathbf{R}^{d}$ maps an image to a class probability vector with $d$ classes. The predicted class for a given sample image $\mathbf{x}$ is argmax$_{i\in C}\mathcal{F}_{\phi}(\mathbf{x})_i$. A targeted adversarial example is generated to mislead the classifier $\mathcal{F}$ into predicting a target label $c\in \mathcal{C}$.

The optimization problem for generating a targeted adversarial example can be formulated as follows:

\begin{equation}
\mathbf{x}_{adv} = \argmin_{|\mathbf{x}_{adv} - \mathbf{x}| \le \epsilon} J(\mathbf{x},c;\phi)
\end{equation}

Some methods like FGSM\cite{goodfellow2015FGSM}, solve this problem by adding perturbation along the direction of the gradient:
\begin{equation}
    \mathbf{x}_{adv} = \operatorname{clip}_{\mathbf{x},\epsilon}\left[\mathbf{x} - \operatorname{sign}(\nabla_{x} J(\mathbf{x},\mathbf{c};\phi))\right].
\end{equation}
where $\operatorname{clip}$ is a function, that clip $\mathbf{x} - \operatorname{sign}(\nabla_{x}\mathbf{J}(\mathbf{x},\mathbf{c};\phi))$ into $\epsilon$-ball of $\mathbf{x}$.


\subsection{Frequency filter}
\label{method:ffa}
Prior research has revealed that CNNs trained on the ImageNet dataset have a strong bias towards the texture and shape of an object~\cite{geirhos2019imagenet}. As illustrated in Fig.~\ref{fig:freq}, the high-frequency component of an image represents its texture, while the low-frequency component represents its shape. To separate the different frequency components of an input image, various methods such as DCT and Fourier transform can be employed. In this study, we use convolution with a low-pass Gaussian filter $W$ as an approximation to obtain the low-frequency part of the image, i.e., $W*\mathbf{x}$. $W$ is a $(4k+1)\times(4k+1)$ kernel matrix, and its formulation is as follows:
\begin{equation}
\label{eq:filter}
    W_{i,j} = \frac{1}{2\pi \sigma^2}\exp{-\frac{i^2+j^2}{2\sigma^2}}
\end{equation}
where $\sigma = k$ denotes the radius of $W$. By using a larger $\sigma$, more high-frequency parts will be filtered. The high-frequency part of an image can be obtained by subtracting its low-frequency part from the original image, i.e., $\mathbf{x} - W * \mathbf{x}$.

Inspired by these findings, several works have been proposed to analyze and attack the vulnerability of deep models from the perspective of frequency, such as Fourier analysis \cite{yin2019fourier}, high-frequency perturbations \cite{wang2020high}, and practical frequency-based attacks \cite{zhang2022practical}. These works show that deep models are vulnerable to perturbations in both high-frequency and low-frequency components. However, there is still a lack of discussion on how to use frequency components to craft targeted class perturbations.

\input{tabs/method_tab}

In Fig.~\ref{fig:freq}, it is evident that image models are highly sensitive to high-frequency components, which makes them capable of predicting these components accurately. To leverage this capability, we attempted to replace the high-frequency components of an image with the high-frequency components of a target class image. Specifically, we used the equation $W*\mathbf{x} + (\mathbf{x}_{target}-W*\mathbf{x}_{target})$ to replace the high-frequency components, as described in Tab.~\ref{tab:method}. We sampled 1000 images from ImageNet~\cite{alex2012ImageNet} and measured the performance of targeted and untargeted attacks after replacing the high-frequency components with those from a 'bull mastiff' image. Our results showed a promising attack success rate for untargeted attacks, but the performance for targeted attacks was unsatisfactory. However, it was still better than several existing methods, such as MI-FGSM. The results described above demonstrate the feasibility of embedding critical frequency information to achieve a targeted attack.

Yin \etal~\cite{yin2019fourier} finds that crafting only high-frequency content in a perturbation results in a small and easily denoised perturbation, which can be classified by adversarially trained models. However, Long \etal~\cite{long2022frequency} suggest addressing the vulnerability of deep models by considering different frequency components. Thus, perturbations should also affect the low-frequency content of an image. Merely replacing the low-frequency content of an image with that of a target class image may cause the resulting image to fall outside the $\epsilon$-ball of the original image $\mathbf{x}$. Conversely, high-frequency components of an image can be considered as small perturbations that can be entirely replaced by a targeted texture. Moreover, to incorporate the targeted information, the low-frequency components must be altered. A generator can be trained to create perturbations based on these two principles.

\subsection{\name}
Taking inspiration from the previous discussion, we suggest a generative method that directly perturbs the low-frequency component of images to create adversarial examples. This perturbation will be added directly to the low-frequency portion of the image to distort low-frequency content and include targeted class information while eliminating the high-frequency information from the original image. Besides, Yang \etal~\cite{yang2022boosting} have pointed out that many existing generative methods train multiple generators for multiple target labels, which is computationally inefficient. They propose a conditional generative approach that enables multi-target class attacks by training only one generator and generating perturbations based on the target class and input images. We hence adopt their generator architecture.

As shown in Fig.~\ref{fig:method}, we train a conditional generator $\mathcal{G}_{\theta}$ with parameters $\theta$ on the entire dataset and class labels. The generator learns a mapping: $(\mathcal{X}, \mathbf{R}^{d}) \rightarrow \mathbf{R}^{H\times W\times C}$. Given sample input image $\mathbf{x}$, and one-hot encoding $\mathcal{I}_{c} \in \mathbf{R}^{d}$ of target class label $c$, $\mathcal{G}_{\theta}$ output adversarial perturbation containing high-frequency texture of given target class image and noise to perturb low-frequency shape. To create an adversarial example, the generated perturbation is directly added to the the low-frequency component of the image using Eq. \ref{eq:filter}, and then projected onto the $\epsilon$-ball of the original image $\mathbf{x}$:
\begin{equation}
\label{eq:pert}
    \mathbf{x}_{adv} = \operatorname{clip}_{\mathbf{x},\epsilon}(\mathcal{G}_{\theta}(\mathbf{x},\mathbbm{1}_{c}) + W*\mathbf{x})
\end{equation}

Given a pretrained network $\mathcal{F}_{\phi}$, and dataset $\mathcal{X}$, the training objective is
\begin{equation}
\label{eq:objective}
    \mathbf{E}_{\substack{\mathbf{x}\sim \mathcal{X}\\ c \sim \mathcal{C}}}[\operatorname{CE}(\mathcal{F}_{\phi}(\operatorname{clip}_{\mathbf{x},\epsilon}(\mathcal{G}_{\theta}(\mathbf{x},\mathbbm{1}_{c}) + W*\mathbf{x})),c)]
\end{equation} 

\input{alg_training}
where $\mathbf{CE}$ is cross-entropy loss. By minimizing this objective, the generator can learn the patterns of target class images that force the classifier to make targeted predictions based on a data distribution. As a result, the generated frequency information is independent of the input image and more generalizable to different models. The optimization procedure is outlined in Algorithm \ref{alg:Train}.

\input{figs/pattern}

Our proposed method, \name, and C-GSP \cite{yang2022boosting} generate adversarial patterns using cross-entropy loss as the loss function. The patterns generated by both methods are structural, repeated, and semantic, with the noise containing information about the target class image. In contrast, methods such as MI-FGSM \cite{goodfellow2015FGSM} generate random noise that lacks semantic information and can easily overfit a specific model. This is why generative models, which learn the pattern of the target class image with respect to the data distribution, often outperform gradient-based methods.

Furthermore, our method focuses more on high-frequency texture and perturbs the low-frequency part of the image, which is more general and less biased. The noise generated by our method is more generalizable to other models and performs better than other generative models, such as CDA.

\section{Experiment}
This section presents our experimental results using the ImageNet dataset~\cite{alex2012ImageNet} to assess the efficacy of our proposed LFAA method for targeted black-box attacks. We describe our experimental setup and implementation details in Section~\ref{setup}, followed by our evaluations of transferability in Section~\ref{targeted_trans}, real-world vision systems in Section~\ref{real_world}, and adversarial defenses in Section~\ref{defense}. We also present an ablation study in Section~\ref{ablation}.

\subsection{Experimental Setup}
\label{setup}

\textbf{Dataset} This paper uses the ImageNet dataset for both training and testing purposes. We trained the generator on 10,000 randomly selected images from the ImageNet training set and evaluated its performance on 1000 images belonging to 1000 categories from the ImageNet validation dataset\cite{alex2012ImageNet}.

\noindent \textbf{Models} We adopt 6 popular models that pre-trained on ImageNet  \ie, ResNet50~\cite{he2016resnet}, Vgg-19$_{BN}$~\cite{Simonyan2015VGG} and DenseNet-121~\cite{huang2017densely}, Inception-v3 (Inc-v3)~\cite{szegedy2016inceptionv3}, Inception-v4 (Inc-v4)~\cite{szegedy2017inceptionv4}, Inception-ResNet-v2(IncRes-v2)~\cite{szegedy2017inceptionv4}. To evaluate the robustness of our attacks against various defense mechanisms, we consider several adversarial defense models including adversarially trained models, denoising defense, and input transformation defense methods. Specifically, we consider the adversarially trained models Inc-v3$_{adv}$ and ensemble adversarially trained network IncRes-v2$_{ens2}$\cite{tramer2018ensemble}. For denoising defense, we consider HGD~\cite{liao2018defense} and NRP~\cite{naseer2020nrp}. For input transformation defense, we consider R\&P~\cite{xie2018mitigating}, NIPS-r3~\footnote{https://github.com/anlthms/nips-2017/tree/master/mmd}, and FD~\cite{liu2019FD}.

\noindent \textbf{Baselines} To evaluate the effectiveness of our proposed \name, we choose several attacks method including MI-FGSM\cite{dong2018boosting}, DIM~\cite{xie2019improving}, TIM~\cite{dong2019evading} and several 
competitive methods on improving the transferability of target attack including Po-Trip \cite{li2020towards} and Logits attack~\cite{zhao2021success}. We also consider several generative approaches \ie, CDA~\cite{naseer2019cross}, and C-GSP\cite{yang2022boosting}.

\noindent\textbf{Implementation Details} In our experiments, we set the maximum perturbation budget $\epsilon$ to be 16. For all the baseline methods, we follow the implementation details specified in their respective papers. Our proposed LFAA is based on ResNet architecture\cite{he2016resnet} and we adopt the same architecture as a previous generative method\cite{yang2022boosting}. Specifically, our generator $\mathcal{G}_{\theta}$ generates perturbations on the low-frequency part of the image with the same input size as the original image. The size of the Gaussian kernel $k$ is set to be 17$\times$17 \ie $k=4$. The classifier $\mathcal{F}{\phi}$ used in our experiments is a standard pre-trained model on ImageNet, and we fixed the parameters $\phi$ in the classifier $\mathcal{F}$ while training the generator $\mathcal{G}{\theta}$. We use the Adam optimizer with a learning rate of 5$\times$10$^{-4}$, $\beta_1 = 0.5$, and $\beta_2=0.999$. All experiments were conducted on a GeForce RTX 3090 GPU using a PyTorch implementation.
\input{tabs/target}
\subsection{Evaluation on Targeted Transferability}
\label{targeted_trans}
We compare the target transferability of our proposed LFAA method with several other attack methods, including MI-FGSM\cite{dong2018boosting}, DIM~\cite{xie2019improving}, TIM~\cite{dong2019evading}, Logits\cite{zhao2021success}, Po+Trip\cite{li2020towards}, CDA\cite{naseer2019cross} and C-GSP\cite{yang2022boosting}. For training-free approaches, we craft adversarial adversaries on three standard trained models and test them on six models. For generative approaches, we train the generator using three standard trained models \ie ResNet50, DenseNet-121, and VGG-19$_{BN}$. For each target label, we generated adversarial examples for each image and evaluated their targeted attack success rate on other models. Fig.~\ref{fig:sample} displays some crafted adversarial examples by LFAA. We randomly sample four target labels from the ImageNet class set, and train generators based on each label. We evaluate the targeted attack success rate, that is the success rate of the victim models to make targeted predictions. The results are summarized in Tab.~\ref{tab:other_model}, each column of this table represents the model to be attacked, while each row indicates the attacker generates the adversarial examples based on the corresponding methods.
\input{figs/tf}

Our experiments show that instance-specific methods \ie MI-FGSM, DIM, TIM, Po-Trip, and Logits, perform well in white-box targeted attacks, and they outperform generation-based attacks in white-box setting. As gradient-based approaches directly update adversarial examples by following the gradient direction from the original image. In contrast, the generative approach learns the semantic patterns from the entire dataset, which may not cover all possible cases encountered during the attack, and hence results in better performance in white-box setting. However, in black-box settings, they tend to overfit specific models, which makes it difficult to transfer the adversarial examples to other models. Generative approaches, including CDA, G-GSP, and our proposed LFAA, have higher black-box targeted attack success rates than the most powerful instance-specific method Logits, with a range from 3.2\% to 76.3\%. Compared to instance-specific methods, the average black-box targeted attack success rate of LFAA is significantly better, with a margin ranging from 14.9\% to 45.5\%. LFAA is also competitive with generative models, with the average targeted success rate of LFAA higher than that of the most powerful generative approach, C-GSP, with a margin ranging from 3.2\% to 15.5\%. These results show that LFAA is superior in generating transferable adversarial examples and support the claim that our approach can significantly improve transferability. 

To further validate the performance of LFAA, we measured the average targeted attack success rate within a family of models. We trained the generator on ResNet50 and Vgg19$_{BN}$ and transferred it to models within the same family (\eg ResNet50 $\rightarrow$ ResNet101, Vgg19$_{BN}$ $\rightarrow$ Vgg16$_{BN}$). The results are shown in Fig.~\ref{fig:family}, our method and C-GSP have similar targeted attack success rates for white-box performance, but for the black-box setting, the performance of our method is better than C-GSP in ResNet family, with a clear margin ranging from 18.7\% to 28.1\%, and for Vgg family, the average targeted attack success rate is higher with a margin ranging from 5.5\% to 34.4\%. This means that for different architectures or within the same architecture family, our method is superior to the state-of-the-art method C-GSP. These results further validate the effectiveness of our proposed LFAA.

\input{tabs/1000_target}
\input{figs/real_world}
\input{figs/exp_pert}
To evaluate LFAA's ability to generate latent representations for targeted labels in the presence of a large label set by using a randomly generated target label dataset. Each image was randomly assigned a target label from a pool of 100 randomly selected labels. To address the challenges posed by large label sets, we follow the approach proposed by Yang \etal~\cite{yang2022boosting} to divide the label set into smaller and more diverse groups. We train a conditional generator for each group and evaluate the transferability of LFAA using two models, each trained on 50 diverse categories. We perform the targeted attack using the corresponding model for the corresponding group when the target label falls into that group. To ensure a fair comparison, we also evaluated our approach against instance-based methods, and due to computing resources, most generative approaches require training multiple generators for multiple labels, which was not feasible in our case, so we only evaluated LFAA against C-GSP. We present the results of our evaluation in Tab.~\ref{tab:100class}, where we summarize the targeted attack success rates of targeted attacks. In the case of instance-based methods, Logits still has the best transferability, but it is weaker than the generative-approach C-GSP. Our LFAA exhibited the strongest transferability, outperforming the most powerful C-GSP with a range of 6.4\% to 21.8\%. Our approach extends the capabilities of C-GSP. Our evaluation demonstrated that LFAA can effectively handle large label sets and further confirmed the superiority of LFAA.
\input{tabs/defense}
\subsection{Evaluation on Real-world Recognition System}
\label{real_world}
The majority of previous works~\cite{guo2019simple,ilyas2018black} have used score-based attacks to fool image recognition systems in the real world, which require thousands of queries to the victim system. In contrast, we conduct an assessment of the effectiveness of targeted transfer attacks by LFAA on the widely-used Google Cloud Vision API. In particular, we generate the adversaries by our conditional generator trained on DenseNet-121 using 100 target classes and transfer the adversarial examples to fool the vision system.

The API provides a list of labels with corresponding scores indicating the confidence of the model for each label. It only returns labels with a score of 50\% or higher and at most 10 labels are shown. Fig.~\ref{fig:real} displays one of the examples where we generated the adversarial examples using the target label "rock snake." The recognition system classified the adversarial images as "snake," which validates that our generator can learn the semantic information of the target class to fool the real-world system.

\subsection{Evaluation on Defense Method}
\label{defense}
To thoroughly evaluate the effectiveness of our proposed method, we assess the attack performance of LFAA against several defense mechanisms, including adversarial training where we consider adversarially trained models Inception-v3 (Inc-v3$_{adv}$), and ensemble adversarially trained network Inception-Resnet-v2($\text{IncRes-v2}_{ens2}$), as well as input-transformation based defenses (R\&P, NIPS-r3, and FD), denoising methods (HGD and NRP). We compared our approach with C-GSP. The results are presented in Tab.~\ref{tab:defense}, where we report the average target attack success rates against each defense method.

The LFAA method performs slightly worse against adversarially trained models, particularly $\text{IncRes-v2}{ens2}$, but has stronger transferability against Inc-v3${adv}$. For input-transformation based defense methods,  LFAA outperforms C-GSP and achieves a higher average target attack success rate with a margin of 15\% against three defense methods. For denoising methods, LFAA has slightly weaker transferability against NRP, but it can still bypass HGD with a higher targeted attack success rate with a margin of 6.6\%. Overall, LFAA has a higher average targeted attack success rate against all seven defense methods than C-GSP, with a margin of 7.4\%. As LFAA contains more semantics regarding the target label, and the underlying pattern still persists under different defense methods. Hence, LFAA can effectively pass some adversarial defense methods. These findings further confirm the effectiveness of our proposed LFAA method.
\input{figs/ablation}
\subsection{Ablation Study}
\label{ablation}
To further gain insight into the performance improvement of LFAA, we conduct ablation and hyper-parameter studies, which train our generator on DenseNet-121 and generate adversarial examples to validate them on the six models.

\textbf{On the effectiveness of Gaussian kernel size $4k+1 \times 4k+1$} To capture the low-frequency components of the image, a Gaussian low-pass filter is utilized for approximation. As shown in Fig.~\ref{fig:ablation}, the performance of transferring adversarial examples into Vgg19$_{BN}$ and Inception-v3 models significantly drops as $k$ increases above 5. On the other hand, when $k$ is less than or equal to 3, the performance of transferring adversarial examples from DenseNet-121 to ResNet50 is not satisfactory. Although $k=4$ shows weak performance in some models, it achieves the highest average targeted attack success rate. Therefore, $k=4$ is chosen as the hyperparameter for the Gaussian low-pass filter. 

\section{Conclusion}
In this paper, we introduce a novel method called LFAA for generating transferable targeted adversarial examples that exploit the vulnerability of deep neural models from a frequency perspective. The proposed approach is capable of generating perturbations that can cause misclassification on multiple black-box target models and real-world vision systems, regardless of the image's source class. LFAA trains a conditional generator to generate targeted adversarial perturbations, which are then added to the low-frequency components of the image. Experimental results on ImageNet demonstrate that LFAA significantly outperforms state-of-the-art methods. This work suggests that different frequency components play a crucial role in deep learning models, and targeted attacks based on perturbing these components can be an effective and efficient approach for generating transferable attacks.

\bibliography{ecai}
\end{document}

%% file: figs/frequency_ex.tex
\begin{figure}
    \centering
    \hspace{0.1cm}
    \begin{minipage}[b]{0.14\textwidth} 
          \centering 
          \includegraphics[width=\linewidth]{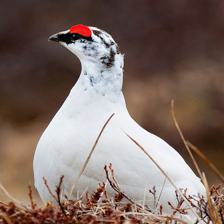}\\
          \vspace{0.3em}
          \includegraphics[width=\linewidth]{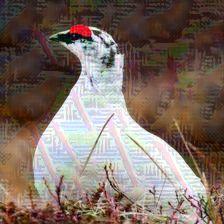}
          \caption*{Raw Image}
    \end{minipage}
    \hspace{0.1cm}
    \begin{minipage}[b]{0.14\textwidth} 
          \centering 
          \includegraphics[width=\linewidth]{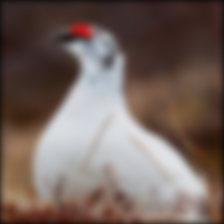}\\
          \vspace{0.3em}
          \includegraphics[width=\linewidth]{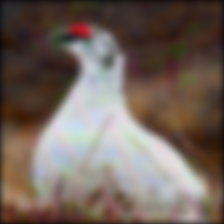}
          \caption*{Low Frequency}
    
    \end{minipage}
    \hspace{0.1cm}
    \begin{minipage}[b]{0.14\textwidth}

          \centering 
          \includegraphics[width=\linewidth]{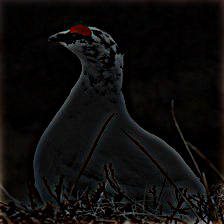}\\
          \vspace{0.3em}
          \includegraphics[width=\linewidth]{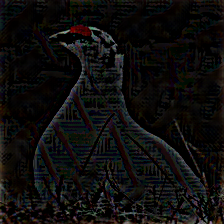}
          \caption*{High Frequency}

    \end{minipage}
    \caption{The frequency component of original image and adversarial images crafted by \name}
    \label{fig:freq}
\end{figure}

%% file: figs/method.tex
\begin{figure*}
    \centering
    \includegraphics[width=\linewidth]{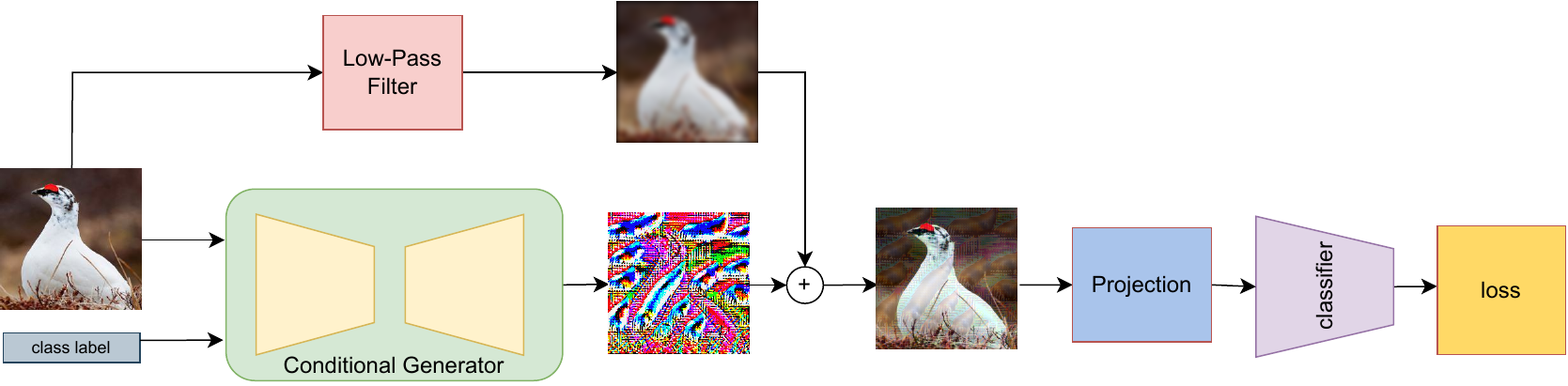}
    \caption{The procedure of our proposed \name}
    \label{fig:method}
\end{figure*}

%% file: tabs/method_tab.tex
\begin{table}[t]
    \begin{center}
    {\caption{Untargeted and targeted attack success rate of several pre-trained models predicting adversarial images with substituted high-frequency component}\label{tab:method}}
    \resizebox{\linewidth}{!}{
    \begin{tabular}{cccccc}
        \toprule
         Metric &ResNet101 & DenseNet121 & Vgg19$_{BN}$ & MobileNet$_{v2}$&Inc-v3\\
         \midrule
         UASR (\%)& 21.4 & 40.1 & 35.4 &41.5 &32.5\\
         TASR (\%)& 6.10 & 0.30 & 1.60 & 0.40 & 0.9\\
         \bottomrule
    \end{tabular}
    }
    \end{center}
\end{table}

%% file: alg_training.tex
\begin{algorithm}[tb]
    \algnewcommand\algorithmicinput{\textbf{Input:}}
    \algnewcommand\Input{\item[\algorithmicinput]}
    \algnewcommand\algorithmicoutput{\textbf{Output:}}
    \algnewcommand\Output{\item[\algorithmicoutput]}

    \caption{\name}
    \label{alg:Train}
	\begin{algorithmic}[1]
		\Input A classifier $\mathcal{F}_{\phi}$ with parameters $\phi$, randomly initialized generative network $\mathcal{G}_{\theta}$, training dataset $\mathbf{X}$, target label $c$ and perturbation budget $\epsilon$.
        \Output adversarial generator $\mathcal{G}_{\theta}$
        \Repeat
        \State Randomly sample batch of images $\mathbf{x} \sim \mathcal{X}$;
        \State Randomly sample batch of class $c \sim \mathcal{C}$;
        \State Forward images $\mathbf{x}$ and one-hot encoding $\mathbbm{1}_{c}$ to Generator $\mathcal{G}_{\theta}$ to obtain the perturbed images $\mathbf{x}_{adv}$ by Eq~\ref{eq:pert};
        \State Forward perturbed images to classifier $\mathcal{F}_{\phi}$ to calculate the loss by Eq\ref{eq:objective};
        \State Backward pass and update parameters of $\mathcal{G}_{\theta}$;
        \Until{$\mathcal{G}_{\theta}$ converge};
        \State \Return $\mathcal{G}_{\theta}$;
	\end{algorithmic} 
\end{algorithm} 

%% file: figs/pattern.tex
\begin{figure}
    \centering
    \hspace{0.1cm}
    \begin{minipage}[c]{0.14\textwidth} 
          \centering 
          \includegraphics[width=\linewidth]{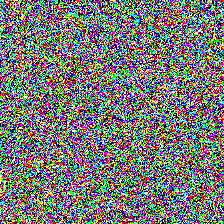}\\
          \vspace{0.3em}
          \includegraphics[width=\linewidth]{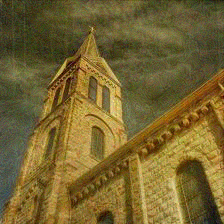}\\
          \caption*{MI-FGSM}
    \end{minipage}
    \hspace{0.1cm}
    \begin{minipage}[c]{0.14\textwidth} 
          \centering 
          \includegraphics[width=\linewidth]{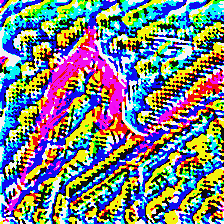}\\
          \vspace{0.3em}
          \includegraphics[width=\linewidth]{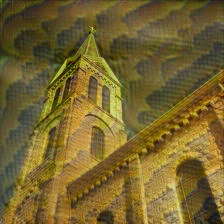}\\
          \caption*{CDA}
    \end{minipage}
    \hspace{0.1cm}
    \begin{minipage}[c]{0.14\textwidth} 
          \centering 
          \includegraphics[width=\linewidth]{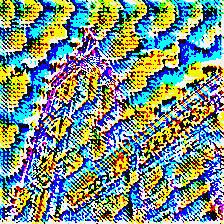}\\
          \vspace{0.3em}
          \includegraphics[width=\linewidth]{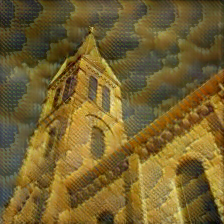}
          \caption*{\name}
    \end{minipage}
    \caption{The Adversarial Perturbation (first row) and Adversarial examples (second row) crafted by different methods}
    \label{fig:pattern}
\end{figure}

%% file: tabs/target.tex
\begin{table*}[t]
\begin{center}
{\caption{Targeted attack success rates (\%) on six models with various adversarial attack. The adversaries are crafted on ResNet-50, DenseNet-121 and Vgg-19$_{BN}$ respectively. * indicates white-box attacks.\label{tab:other_model}}}
\scalebox{1.1}{
\begin{tabular}{c>{\rowmac}c>{\rowmac}c>{\rowmac}c>{\rowmac}c>{\rowmac}c>{\rowmac}c>{\rowmac}c>{\rowmac}c}
\toprule
Model & Attack & ResNet-50  & DenseNet-121 & Vgg19$_{BN}$ &Inc-v3 & Inc-v4 & IncRes-v2\\
\midrule
\multirow{8}{*}{ResNet-50}& MI-FGSM & ~~98.4*  & ~~0.4 & ~~0.2 & ~~0.1 & ~~0.1 &  ~~0.1\\
& DIM & ~~84.5*  & ~2.4 & ~~1.1 & ~~0.8 & ~~0.2 & ~~0.1\\
& TIM & ~~99.9*  & ~1.3 & ~~0.8 & ~~0.5 & ~~0.4 & ~~0.5\\
& Po-Trip & \textbf{100.0}*  & ~~3.8 & ~~2.5 & ~~1.0 & ~~0.2 & ~~0.6\\
& Logits & ~~98.1*  & ~~9.8 & ~~3.9 & ~~1.5 & ~~0.9& ~~2.5\\
& CDA & ~~77.3*  & 29.2 & 35.1 & ~~4.2 & ~~7.0 & ~~2.8  \\
& C-GSP & ~~92.4*  & 55.2 & 36.2 & 29.9 & 24.3 & 11.5\\
& \name & ~~95.8*  & \textbf{80.1} & \textbf{66.6}  &\textbf{38.9} & \textbf{33.8} & \textbf{15.2}& \\
\midrule
\multirow{8}{*}{DenseNet121}& MI-FGSM & ~~0.8 & \textbf{100.0}* & 
~~0.5 & ~~0.5 & ~~0.2 &  ~~0.3\\
& DIM & ~~2.6 & ~~89.4* & ~~2.6 & ~~1.8& ~~0.5 & ~~0.1\\
& TIM & ~~1.0 & ~\textbf{100.0}* & ~~0.4 & ~~0.3 & ~~0.5 & ~~0.3\\
& Po-Trip & ~~2.6  & \textbf{100.0}* & ~~1.1 & ~~1.0 & ~~0.4 & ~~0.4\\
& Logits & ~~4.3 & \textbf{100.0}* & ~~2.6 & ~~2.5 & ~~1.2 & ~~1.1\\
& CDA & 55.6  &~~89.5* & 27.3 & 22.3 & ~~8.7 & ~~2.4 \\
& C-GSP & 51.7  & ~~92.7* & 33.6 & 29.6 & 17.9 & 12.1\\
& \name & \textbf{72.9}  &~~94.5*& \textbf{48.7} & \textbf{33.2} &\textbf{39.2} & \textbf{26.0} \\
\midrule
\multirow{8}{*}{Vgg19$_{BN}$}& MI-FGSM & ~~0.4 & ~~0.3 & ~~99.9* & ~~0.2 & ~~0.2 &  ~~0.2\\
& DIM & ~~0.7  & ~~0.7 & ~~83.4* & ~~0.4 & ~~0.3 & ~~0.2\\
& TIM & ~~0.4  & ~~0.4 & \textbf{100.0}* & ~~0.2 & ~~0.1 & ~~0.6\\
& Po-Trip & ~~0.7  & ~~0.7 & \textbf{100.0}* & ~~0.6 & ~~0.3 & ~~0.3\\
& Logits & ~~1.8  & ~~2.2 & \textbf{100.0}* & ~~0.7 & ~~1.0 & ~~0.9\\
& CDA & 10.3  & 12.3 & ~~96.3* & ~~0.7 & ~~1.1 & ~~0.1 \\
& C-GSP & 19.7  & 22.6 & ~~92.0* & ~~8.2 & \textbf{11.7} & ~~1.3\\
& \name & \textbf{29.6}  & \textbf{29.3} & ~~93.7* & \textbf{11.2} & ~~7.7 & ~~\textbf{1.8}\\
\bottomrule
\end{tabular}
}
\end{center}
\end{table*}

%% file: figs/tf.tex
\begin{figure}[t]
    \centering
    \hspace{0.1cm}
    \begin{minipage}[c]{0.2\textwidth} 
          \centering 
          \includegraphics[width=\linewidth]{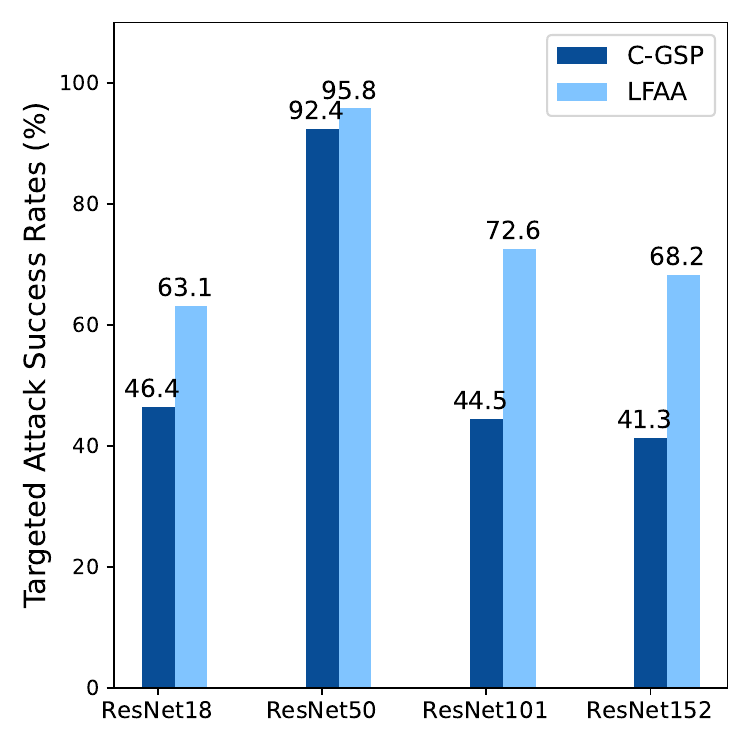}\\
          \caption*{ResNet Family}
    \end{minipage}
    \hspace{0.2cm}
    \begin{minipage}[c]{0.2\textwidth} 
          \centering 
          \includegraphics[width=\linewidth]{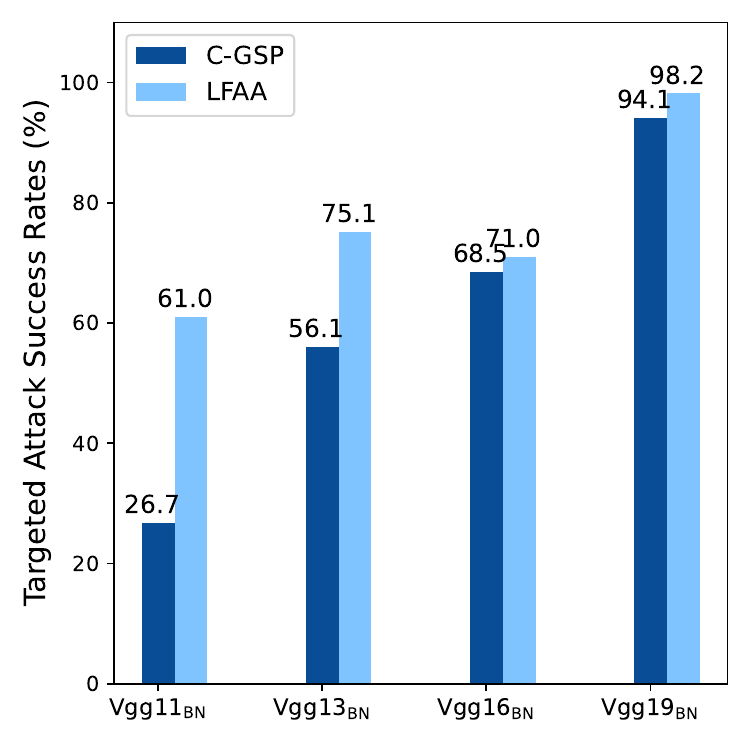}\\
          \caption*{VGG Family}
    \end{minipage}
    \hspace{0.1cm}
    \caption{Targeted attack success rates (\%) on eight models by G-GSP and LFAA, the generators are trained on ResNet-50, and Vgg19$_{BN}$.}
    \label{fig:family}
\end{figure}

%% file: tabs/1000_target.tex
\begin{table}[t]
\begin{center}
{\caption{Target attack success rates (\%) on six models under single model setting with various attack methods. The adversaries are crafted on DenseNet-121.}\label{tab:100class}}
\begin{tabular}{cccccc}
\toprule
 Attack   &ResNet50 & Vgg19$_{BN}$ &Inc-v3 & Inc-v4 & IncRes-v2\\
\midrule
MI-FGSM  & ~~2.5 & ~~1.5 & ~~0.9 & ~~1.3 &  ~~2.0\\
 DIM &~0.8 & ~~0.6 & ~~0.6 & ~~0.3 & ~~0.6\\
 TIM&~1.0 & ~~0.4 & ~~0.3 & ~~0.5 & ~~0.3\\
 Po-Trip   & ~~2.5 & ~~1.5 & ~~0.9 & ~~1.3 &  ~~2.0\\
Logits  & ~~5.8 & ~~3.6 & ~~1.8 & ~~1.6& ~~2.7\\
 C-GSP  & 18.0 & 16.3 & 5.2 & ~~3.6 & ~~2.8\\
 \name   & \textbf{39.8} & \textbf{31.0}  &\textbf{18.0} & \textbf{10.9} & ~~\textbf{9.2}\\
\bottomrule
\end{tabular}

\end{center}
\vspace{-1.5em}
\end{table}

%% file: figs/real_world.tex
\begin{figure}[t]
    \centering
    \hspace{0.1cm}
    \begin{minipage}[c]{0.21\textwidth} 
          \centering 
          \includegraphics[width=\linewidth]{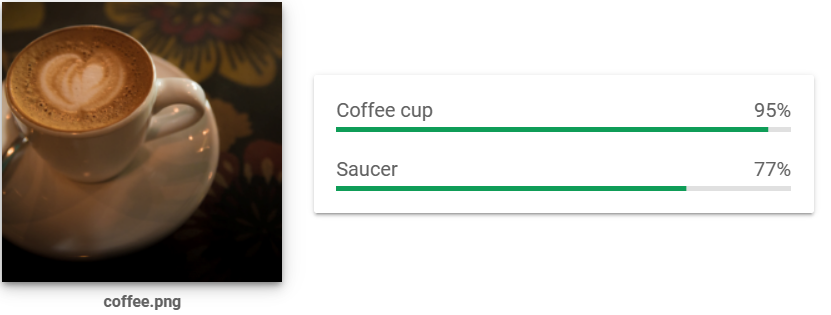}\\
          \vspace{0.3em}
          \caption*{Original}
    \end{minipage}
    \hspace{0.1cm}
    \begin{minipage}[c]{0.21\textwidth} 
          \centering 
          \includegraphics[width=\linewidth]{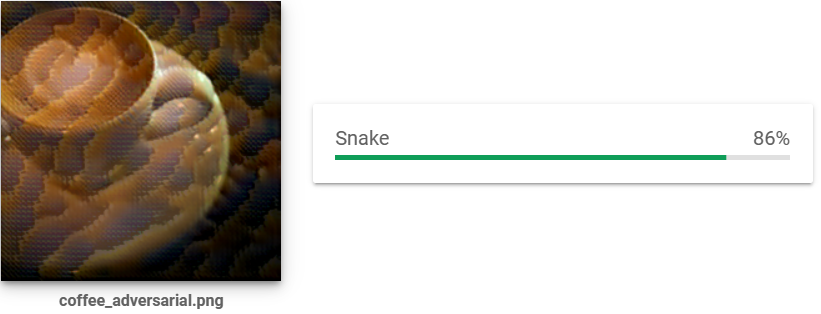}\\
          \vspace{0.3em}
          \caption*{Adversarial}
    \end{minipage}
    \caption{Successful targeted adversarial images on Google
Cloud Vision generated by \name, the given target class is rock snake.}
    \label{fig:real}
     \vspace{-0.4em}
\end{figure}

%% file: figs/exp_pert.tex
\begin{figure*}
    \centering
    \begin{minipage}[b]{0.14\textwidth} 

          \centering 
          \caption*{kite}
          \includegraphics[width=\linewidth]{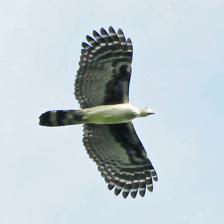}\\
          \vspace{0.3em}
          \includegraphics[width=\linewidth]{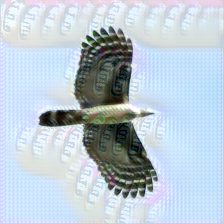}
          \caption*{combination lock}

    \end{minipage}
    \hspace{0.1cm}
    \begin{minipage}[b]{0.14\textwidth} 

          \centering 
          \caption*{Mergus serrator}
          \includegraphics[width=\linewidth]{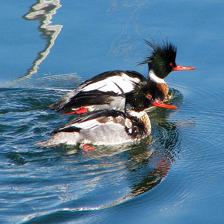}\\
          \vspace{0.3em}
          \includegraphics[width=\linewidth]{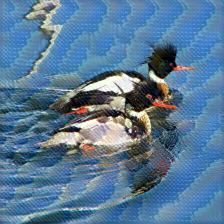}
          \caption*{rock snake}

    \end{minipage}
    \hspace{0.1cm}
    \begin{minipage}[b]{0.14\textwidth} 

          \centering 
          \caption*{dock}
          \includegraphics[width=\linewidth]{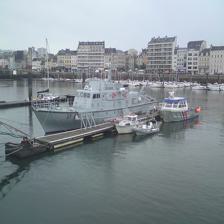}\\
          \vspace{0.3em}
          \includegraphics[width=\linewidth]{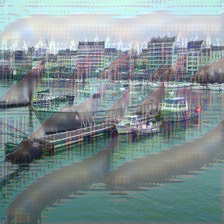}
          \caption*{sea lion}


    \end{minipage}
    \hspace{0.1cm}
    \begin{minipage}[b]{0.14\textwidth} 

          \centering 
          \caption*{fireboat}
          \includegraphics[width=\linewidth]{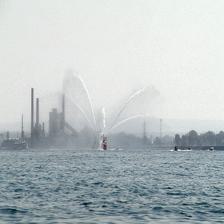}\\
          \vspace{0.3em}
          \includegraphics[width=\linewidth]{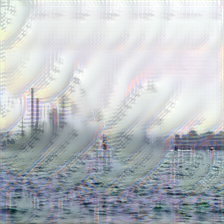}
          \caption*{barometer}

    \end{minipage}
     \hspace{0.1cm}
    \begin{minipage}[b]{0.14\textwidth} 
         \centering 
          \caption*{viaduct}
          \includegraphics[width=\linewidth]{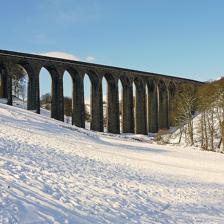}\\
          \vspace{0.3em}
          \includegraphics[width=\linewidth]
          {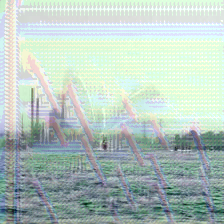}
          \caption*{swing}

    \end{minipage}
    \caption{Targeted adversaries produced by generator trained against ResNet50. 1st row shows original images, while 2nd row shows adversaries examples. The caption in the top represents the
original category. The caption below represents the target category. }
    \label{fig:sample}
\end{figure*}

%% file: tabs/defense.tex
\begin{table*}[t]
\begin{center}
{\caption{Target attack success rates (\%) of seven defense methods by C-GSP and \name. The generators are trained on ResNet-50\label{tab:defense}}}
\scalebox{1.15}{
\begin{tabular}{lccccccccc}
\toprule
Method & Inc-v3$_{adv}$ &IncRes-v2$_{ens}$& HGD & R\&P & NIPS-r3 & FD    & NRP & Average\\
\midrule
C-GSP & 15.9 & \textbf{23.5} & 62.0 & 38.3           &  37.1     & 16.0             &\textbf{13.2}         & 29.4\\
\name & \textbf{24.2} &18.2  & \textbf{68.6} & \textbf{57.5} & \textbf{57.1} & \textbf{21.8} & 10.1 & \textbf{36.8}\\
\bottomrule
\end{tabular}
}
\end{center}
\end{table*}

%% file: figs/ablation.tex
\begin{figure}
    \centering
    \includegraphics[width=0.6\linewidth]{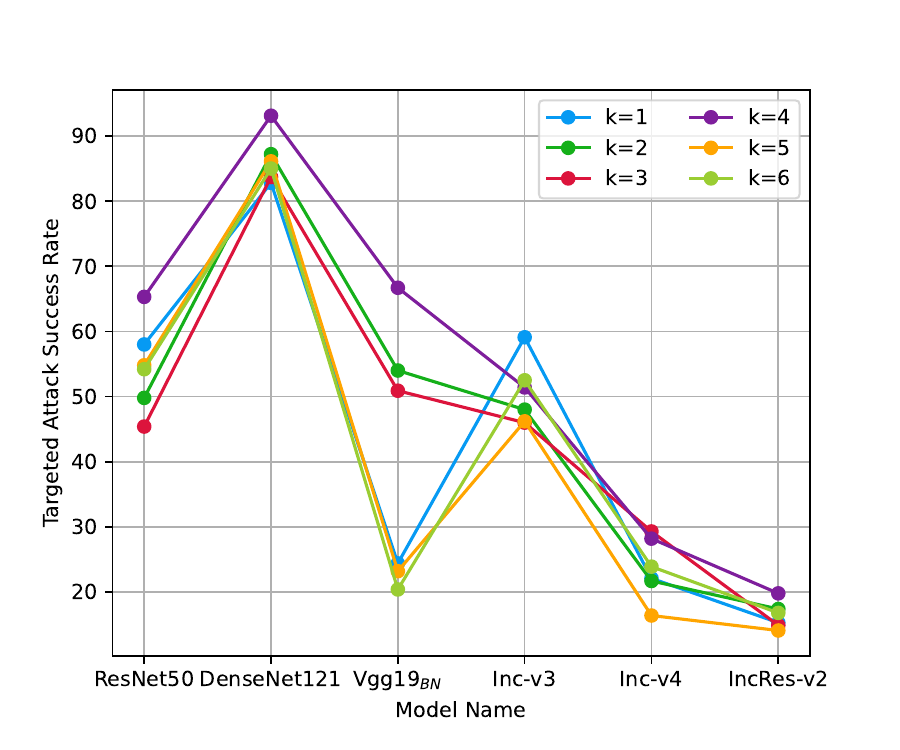}
    \caption{Targeted attack success rates (\%) on six models by LFAA with different $k$.}
    \label{fig:ablation}
\end{figure}

%% file: ecai-sample-and-instructions.bbl
\newcommand{\arxiv}{arXiv preprint arXiv:}
\begin{thebibliography}{10}

\bibitem{brendel2018decision}
Wieland Brendel, Jonas Rauber, and Matthias Bethge, `{Decision-Based
  Adversarial Attacks: Reliable Attacks Against Black-Box Machine Learning
  Models}', {\em International Conference on Learning Representations}, (2018).

\bibitem{dong2018boosting}
Yinpeng Dong, Fangzhou Liao, Tianyu Pang, Hang Su, Jun Zhu, Xiaolin Hu, and
  Jianguo Li, `{Boosting Adversarial Attacks with Momentum}', in {\em
  Proceedings of the IEEE Conference on Computer Vision and Pattern
  Recognition}, pp. 9185--9193, (2018).

\bibitem{dong2019evading}
Yinpeng Dong, Tianyu Pang, Hang Su, and Jun Zhu, `{Evading Defenses to
  Transferable Adversarial Examples by Translation-invariant Attacks}', in {\em
  Proceedings of the IEEE/CVF Conference on Computer Vision and Pattern
  Recognition}, pp. 4312--4321, (2019).

\bibitem{gao2021feature}
Lianli Gao, Yaya Cheng, Qilong Zhang, Xing Xu, and Jingkuan Song, `{Feature
  Space Targeted Attacks by Statistic Alignment}', in {\em Proceedings of the
  International Joint Conference on Artificial Intelligence}, pp. 671--677,
  (2021).

\bibitem{geirhos2019imagenet}
Robert Geirhos, Patricia Rubisch, Claudio Michaelis, Matthias Bethge, Felix~A.
  Wichmann, and Wieland Brendel, `{Imagenet-trained cnns are biased towards
  texture; increasing shape bias improves accuracy and robustness}', {\em
  International Conference on Learning Representations}, (2019).

\bibitem{goodfellow2015FGSM}
Ian~J Goodfellow, Jonathon Shlens, and Christian Szegedy, `{Explaining and
  harnessing adversarial examples}', in {\em International Conference on
  Learning Representations}, (2015).

\bibitem{guo2019low}
Chuan Guo, Jared~S. Frank, and Kilian~Q. Weinberger, `{Low Frequency
  Adversarial Perturbation }', in {\em Proceedings of the Thirty-Fifth
  Conference on Uncertainty in Artificial Intelligence}, volume 115, pp.
  1127--1137, (2019).

\bibitem{guo2019simple}
Chuan Guo, Jacob Gardner, Yurong You, Andrew~Gordon Wilson, and Kilian
  Weinberger, `{Simple Black-box Adversarial Attacks}', in {\em International
  Conference on Machine Learning}, pp. 2484--2493, (2019).

\bibitem{he2016resnet}
Kaiming He, Xiangyu Zhang, Shaoqing Ren, and Jian Sun, `{Deep Residual Learning
  for Image Recognition}', in {\em Proceedings of the IEEE Conference on
  Computer Vision and Pattern Recognition}, pp. 770--778, (2016).

\bibitem{huang2017densely}
Gao Huang, Zhuang Liu, Laurens Van Der~Maaten, and Kilian~Q Weinberger,
  `{Densely connected convolutional networks}', in {\em Proceedings of the IEEE
  Conference on Computer Vision and Pattern Recognition}, pp. 4700--4708,
  (2017).

\bibitem{ilyas2018black}
Andrew Ilyas, Logan Engstrom, Anish Athalye, and Jessy Lin, `{Black-box
  Adversarial Attacks with Limited Queries and Information}', in {\em
  International Conference on Machine Learning}, pp. 2142--2151, (2018).

\bibitem{Simonyan2015VGG}
Simonyan Karen and Zisserman Andrew, `{Very Deep Convolutional Networks for
  Large-Scale Image Recognition}', in {\em International Conference on Learning
  Representations}, (2015).

\bibitem{alex2012ImageNet}
Alex Krizhevsky, Ilya Sutskever, and Geoffrey~E. Hinton, `{ImageNet
  Classification with Deep Convolutional Neural Networks}', in {\em Advances in
  Neural Information Processing Systems}, pp. 1106--1114, (2012).

\bibitem{li2020towards}
Maosen Li, Cheng Deng, Tengjiao Li, Junchi Yan, Xinbo Gao, and Heng Huang,
  `{Towards Transferable Targeted Attack}', in {\em Proceedings of the IEEE
  Conference on Computer Vision and Pattern Recognition}, pp. 638--646, (2020).

\bibitem{liao2018defense}
Fangzhou Liao, Ming Liang, Yinpeng Dong, Tianyu Pang, Xiaolin Hu, and Jun Zhu,
  `{Defense against Adversarial Attacks using High-level Representation Guided
  Denoiser}', in {\em Proceedings of the IEEE/CVF Conference on Computer Vision
  and Pattern Recognition}, pp. 1778--1787, (2018).

\bibitem{lin2020nesterov}
Jiadong Lin, Chuanbiao Song, Kun He, Liwei Wang, and John~E. Hopcroft,
  `{Nesterov Accelerated Gradient and Scale Invariance for Adversarial
  Attacks}', in {\em International Conference on Learning Representations},
  (2020).

\bibitem{liu2019FD}
Zihao Liu, Qi~Liu, Tao Liu, Nuo Xu, Xue Liu, Yanzhi Wang, and Wujie Wen,
  `{Feature distillation: Dnn-oriented jpeg compresstion against adversarial
  examples}', in {\em Proceedings of the IEEE/CVF Conference on Computer Vision
  and Pattern Recognition}, pp. 860--868, (2019).

\bibitem{long2022frequency}
Yuyang Long, Qilong Zhang, Boheng Zeng, Lianli Gao, Xianglong Liu, Jian Zhang,
  and Jingkuan Song, `{Frequency Domain Model Augmentation for Adversarial
  Attack}', {\em European Conference on Computer Vision}, (2022).

\bibitem{madry2018pgd}
Aleksander Madry, Aleksandar Makelov, Ludwig Schmidt, Dimitris Tsipras, and
  Adrian Vladu, `{Towards Deep Learning Models Resistant to Adversarial
  Attacks}', in {\em International Conference on Learning Representations},
  (2018).

\bibitem{Moosavi2017universal}
Seyed-Mohsen Moosavi-Dezfooli, Alhussein Fawzi, Omar Fawzi, and Pascal
  Frossard, `{Universal Adversarial Perturbations}', in {\em Proceedings of the
  IEEE Conference on Computer Vision and Pattern Recognition}, pp. 1765--1773,
  (2017).

\bibitem{naseer2019cross}
Muhammad~Muzammal Naseer, Salman~H Khan, Muhammad~Haris Khan, Fahad
  Shahbaz~Khan, and Fatih Porikli, `{Cross-domain transferability of
  adversarial perturbations}', volume~32, pp. 12905--12915, (2019).

\bibitem{naseer2020nrp}
Muzammal Naseer, Salman Khan, Munawar Hayat, Khan~Fahad Shahbaz, and Fatih
  Porikli, `{A Self-supervised Approach for Adversarial Robustness}', in {\em
  Proceedings of the IEEE/CVF Conference on Computer Vision and Pattern
  Recognition}, pp. 262--271, (2020).

\bibitem{Papernot2017blackbox}
Nicolas Papernot, Patrick McDaniel, Ian Goodfellow, Somesh Jha, Z~Berkay Celik,
  and Ananthram Swami, `{Practical black-box attacks against machine
  learning}', in {\em Proceedings of the 2017 ACM on Asia conference on
  computer and communications security}, pp. 506--519, (2017).

\bibitem{poursaeed2018generative}
Omid Poursaeed, Isay Katsman, Bicheng Gao, and Serge Belongie, `{Generative
  adversarial perturbations}', in {\em Proceedings of the IEEE Conference on
  Computer Vision and Pattern Recognition}, pp. 4422--4431, (2018).

\bibitem{sharma2019on}
Yash Sharma, Gavin~Weiguang Ding, and Marcus~A. Brubaker, `{On the
  effectiveness of low frequency perturbations }', in {\em Proceedings of the
  International Joint Conference on Artificial Intelligence}, p. 3389–3396,
  (2019).

\bibitem{szegedy2017inceptionv4}
Christian Szegedy, Sergey Ioffe, Vincent Vanhoucke, and Alex Alemi,
  `{Inception-v4, inception-resnet and the impact of residual connections on
  learning}', in {\em AAAI Conference on Artificial Intelligence}, (2017).

\bibitem{szegedy2016inceptionv3}
Christian Szegedy, Vincent Vanhoucke, Sergey Ioffe, Jon Shlens, and Wojna
  Zbjgniew, `{Rethinking the inception architecture for computer vision}', in
  {\em Proceedings of the IEEE Conference on Computer Vision and Pattern
  Recognition}, pp. 2818--2826, (2016).

\bibitem{szegedy2014intriguing}
Christian Szegedy, Wojciech Zaremba, Ilya Sutskever, Joan Bruna, Dumitru Erhan,
  Ian Goodfellow, and Rob Fergus, `{Intriguing properties of neural networks}',
  in {\em International Conference on Learning Representations}, (2014).

\bibitem{tramer2018ensemble}
Florian Tram{\`e}r, Alexey Kurakin, Nicolas Papernot, Ian Goodfellow, Dan
  Boneh, and Patrick McDaniel, `{Ensemble Adversarial Training: Attacks and
  Defenses}', {\em International Conference on Learning Representations},
  (2018).

\bibitem{wang2021admix}
Xiaosen Wang, Xuanran He, Jingdong Wang, and Kun He, `Admix: Enhancing the
  transferability of adversarial attacks', in {\em International Conference on
  Computer Vision}, pp. 16138--16147, (2021).

\bibitem{wang2020high}
Xindi Wang, Haohan an~Wu, Zeyi Huang, and Eric~P. Xing, `{High-Frequency
  Component Helps Explain the Generalization of Convolutional Neural
  Networks}', in {\em Proceedings of the IEEE Conference on Computer Vision and
  Pattern Recognition}, pp. 8684--8694, (2020).

\bibitem{wang2021Feature}
Zhibo Wang, Hengchang Guo, Zhifei Zhang, Wenxin Liu, Qin Zhan, and Kui Ren,
  `{Feature importance-aware transferable adversarial attacks}', in {\em
  International Conference on Computer Vision}, pp. 7639--7648, (2021).

\bibitem{wu2020skip}
Dongxian Wu, Yisen Wang, Shu-Tao Xia, James Bailey, and Xingjun Ma, `{Skip
  Connections Matter: On the Transferability of Adversarial Examples Generated
  with ResNets}', in {\em Proceedings of the International Conference on
  Learning Representations}, (International Conference on Learning
  Representations).

\bibitem{xie2018mitigating}
Cihang Xie, Jianyu Wang, Zhishuai Zhang, Zhou Ren, and Alan Yuille,
  `{Mitigating adversarial effects through randomization}', in {\em
  International Conference on Learning Representations}, (2018).

\bibitem{xie2019improving}
Cihang Xie, Zhishuai Zhang, Yuyin Zhou, Song Bai, Jianyu Wang, Zhou Ren, and
  Alan~L Yuille, `{Improving Transferability of Adversarial Examples with Input
  Diversity}', in {\em Proceedings of the IEEE Conference on Computer Vision
  and Pattern Recognition}, pp. 2730--2739, (2019).

\bibitem{yang2022boosting}
Xiao Yang, Yinpeng Dong, Tianyu Pang, Hang Su, and Jun Zhu, `{Boosting
  Transferability of Targeted Adversarial Examples via Hierarchical Generative
  Network}', in {\em Proceedings of the European Conference on Computer Vision
  (ECCV)}, p. 725–742, (2022).

\bibitem{yin2019fourier}
Dong Yin, Raphael~Gontijo Lopes, Jonathon Shlens, Ekin~D. Cubuk, and Justin
  Gilmer, `{A fourier perspective on model robustness in computer vision}',
  volume 1189, p. 13276–13286, (2019).

\bibitem{zhang2022practical}
Qilong Zhang, Chaoning Zhang, Chaoqun Li, Jingkuan Song, and Lianli Gao,
  `{Practical No-box Adversarial Attacks with Training-free Hybrid Image
  Transformation}', {\em \arxiv{}2203.04607}, (2022).

\bibitem{zhao2021success}
Zhengyu Zhao, Zhuoran Liu, and Martha Larson, `{On Success and Simplicity: A
  Second Look at Transferable Targeted Attacks}', in {\em Advances in Neural
  Information Processing Systems}, pp. 6115--6128, (2021).

\end{thebibliography}
